# Towards a neural architecture of language: Deep learning versus logistics of access in neural architectures for compositional processing

*Frank van der Velde*

Cognition, Data and Education,
University of Twente, The Netherlands
f.vandervelde@utwente.nl
veldefvander@outlook.com

**Abstract**

Recently, a number of articles have argued that deep learning models such as GPT could also capture key aspects of language processing in the human mind and brain. However, I will argue that these models are not suitable as neural models of human language. Firstly, because they fail on fundamental boundary conditions, such as the amount of learning they require. This would in fact imply that the mechanisms of GPT and brain language processing are fundamentally different. Secondly, because they do not possess the logistics of access needed for compositional and productive human language processing. Neural architectures could possess logistics of access based on small-world like network structures, in which processing does not consist of symbol manipulation but of controlling the flow of activation. In this view, two complementary approaches would be needed to investigate the relation between brain and cognition. Investigating learning methods could reveal how 'learned cognition' as found in deep learning could develop in the brain. However, neural architectures with logistics of access should also be developed to account for 'productive cognition' as required for natural or artificial human language processing. Later on, these approaches could perhaps be combined to see how such architectures could develop by learning and development from a simpler basis.

## Table of Contents

## 1. Introduction

Tremendous progress has been made with deep learning in neural networks, as exemplified with pattern recognition (e.g., LeCun et al., 2015), gaming (e.g., Silver et al. 2017) and language (e.g., Brown et al., 2020). Regardless of the way in which the networks are trained or how many parameters or training data (e.g., sentences) are needed, it is remarkable to see that structural forms of knowledge, way above mere associations, can be embedded in and retrieved from neural networks, based on learning

This, in itself, is an important discovery for understanding the brain. The connections in the brain will be modified by experience throughout life. The actual forms of learning (or development) may be different, but deep learning indicates that structural knowledge could be acquired in this way.

Recently, a number of articles (e.g., Schrimpf et al., 2021; Caucheteux et al., 2022) have argued that deep learning models such as GPT could also capture key aspects of language processing in the human mind and brain. For example, Schrimpf et al. (2021) showed that a number of deep learning models could also predict brain activity observed during language processing. In particular, the models that were better in prediction the next word in a sentence were also better in predicting neural activity observed in sentence processing (Blank et al., 2014; Fedorenko et al, 2016; Pereira et al., 2018).

The success of these AI models in predicting brain activity could suggest that predicting the next word in a sentence, as performed by the model, is also important in human sentence processing. This is indeed an important observation, in line with suggestions on the role of predictions in language processing (e.g., Kuperberg & Jaeger, 2016; Goldstein et al., 2021).

Moreover, it could also imply that these AI models capture fundamental computational aspects of brain processing. Schrimpf et al. (2021, p. 2b) express this idea as follows: "though the goal of contemporary AI is to improve model performance and not necessarily to build models of brain processing, this endeavor appears to be rapidly converging on architectures that might capture key aspects of language processing in the human mind and brain."

However, here I will argue that AI models in their current form fail as models of human language processing on two main aspects. Firstly, they fail to meet fundamental boundary conditions that such models would have to fulfill. This in fact suggests that their mechanisms and those of the brain are fundamentally different. Secondly, they do not provide the productivity of compositional processing that underlies human language, specifically because they do not provide the 'logistics of access' needed for this. The focus in this discussion will be on GPT models, which also performed the best in Schrimpf et al. (2021).

On the basis of this discussion I will argue that an integration of forms of learning with more explicit or designed architectures aimed at providing logistics of access would be needed to understand the neural basis of human language, and develop AI models for productive language processing.

The outline of this article is as follows:

The next section discusses why GPT models fail to meet fundamental boundary conditions for models of human language processing, and what this implies for understanding brain mechanisms.

Section 3 discusses the lack of productivity of GPT, with an emphasis on the importance of 'logistics of access' in human language processing (and cognition in general).

Section 4 discusses the nature of logistics of access and why it is the basis of compositional and productive processing as found in human language and cognition.

Section 5 discusses a classical example of logistics of access based on the use of symbols and processing with symbol manipulation.

Section 6 briefly discusses a neural architecture in which logistics of access results from small-world like network structures and processing consists of controlling the flow of activation instead of symbol manipulation.

The final section briefly discusses the integration of 'learned cognition', given by learning methods as found in deep learning, and 'productive cognition' based on neural architectures designed to possess logistics of access.

## 2. GPT as models of brain mechanisms: failing on boundary conditions

To analyse the role of GPT models as potential models of brain mechanisms underlying human language, in particular sentence processing, we have to look at the way these models are trained. Typically, there are two stages of training of GPT models (Brown et al., 2020). In the first stage, often referred to as pre-training, the models are trained on huge sets of sentences in an unsupervised manner. In the second stage, the model can then be trained for a particular task in a supervised manner. Schrimpf et al. (2021) also applied and tested this second learning stage.

In the pre-training stage, GPT models are trained on more sentences than a human will see in a lifetime (Brown et al., 2020). For engineering purposes, this would be fine (although the energy costs related to it are staggering, e.g., Brown et al., 2020; see also García-Martín et al., 2019). But as neural models of human sentence processing this violates a simple but fundamental boundary condition: humans cannot learn more sentences than they will see in a lifetime.

This touches upon an important difference between engineering and science. In engineering one can apply anything that would lead to a desired result. But a scientific model or theory has to adhere to 'boundary conditions' imposed by the domain at hand. Physics provides ample examples of these. Here, pre-training models using more sentences than humans will see in a lifetime is fine from an engineering perspective. But it would disqualify such models as scientific models of language processing in the human mind and brain.

However, one could argue that such a pre-training stage is needed and allowed to give the model the abilities that evolution has provided for human language processing. As far as we know, the human brain is the only nervous system capable of processing and producing a language at the complexity level of natural language (e.g., Hauser et al., 2002). For example, the failure of chimpanzees or bonobos to learn natural language (e.g., Terrace, 2019) suggests that the brains of these animals do not possess the neural structures that are required for successful language learning. Viewed in this way, the pre-training stage of GPT would perhaps be needed to acquire these neural structures. If so, then using a huge set of sentences would not be a problem, as it would be a way to get on a par with a long evolutionary development.

But the pre-training stage of GPT is language specific. Doing it, e.g., for English will not prepare the model to deal with any other language[1]. In contrast, the ability for language that humans acquired during evolution is not language specific. A child will learn any language it is exposed to. So, the huge set of sentences used in pre-training GPT models violates a fundamental boundary condition on language processing in the human mind and brain in any way you look at it.

Further problems with GPT learning as a scientific model of human neural language processing in fact emerged from the study of Schrimpf et al. (2021) itself. They varied the amount of training data and the size of the vocabulary in the second training stage and observed a strong negative correlation between increase in training data and vocabulary and the performance of the GPT model (and other models). Again, this seems to violate an important boundary condition on human language learning. Humans learn language in an incremental manner, starting with simple sentences and a small vocabulary and gradually building that up to adult size (e.g., Saxton, 2010). During this process, their language abilities improve, not decline. Any model of human language processing should exhibit the same kind of behavior. That is, it would have to learn language incrementally. Any other way would violate another important boundary condition on human language processing.

The negative correlation between the amount of learning and performance of the models studied by Schrimpf et al. (2021) is perhaps not so surprising when viewed from the perspective of how these models learn (van der Velde, 2015a), as exemplified in the phenomenon of catastrophic interference, in which newly learned material overrides the ability to process previously learned material.

## 2.1 The same results with different mechanisms

At present, deep learning models still fail on a number of basic language tasks, such as question answering (e.g., Krishna et al., 2021). But even if their performance eventually is on a par with that of humans, could we conclude from this that their mechanisms are similar to those of the brain?

As analyzed above, GPT models learn language in a language specific way, using a very large set of sentence for training. In contrast, humans learn an initially arbitrary language in an incremental way, using a limited set of training sentences. So, even if the end results would be the same, the conditions under which these are obtained are fundamentally different.

This would imply that the underlying mechanisms are fundamentally different. For example, you could travel 1000 miles with a bicycle in days or with a racecar in hours. In this case the underlying mechanisms are clearly very different, even though the end result is the same. It is also clear that this is indicated by the different conditions under which the same result is obtained. So, these should be taken into account in analysing similarities between underlying mechanisms.

Furthermore, the simulation of brain activity with GPT is indirect. For example, Schrimpf et al. (2021) simulated brain activity by first fitting a linear regression model between GPT activity and brain activity, using 80% of the available data. Then, this regression model could predict the remaining brain activity data on the basis of the remaining GPT data.

This indeed suggests a similarity between word prediction with GPT and word prediction in the brain. Based on its learning behavior, 80% of the input of a GPT model would predict its behavior on the remaining input data, as this follows from the general characteristics of function approximation with neural networks (van der Velde, 2015a). The success of the regression model illustrates that 80% of brain activity is predictive of the remaining brain activity as well, which could indicate that the underlying brain processes are also related to word prediction by means of pattern recognition. However, this does not preclude the possibility that these mechanisms are different from those of GPT. The conclusion that they are similar would require more direct simulations of brain activity (e.g., see section 6.1).

Moreover, GPT models are trained with semantic information generated by humans, i.e. human brains. So, a linear regression between GPT and brain data could be seen as a linear regression between a statistical compression of human semantic data and semantic representation in the brain. This is indeed very interesting, as it could suggest that semantic representation in the brain follows generic patterns, as e.g. observed by Huth et al. (2016).

However, this is still a far cry from concluding that GPT language processing is similar to that of the brain, as further analyzed below.

### 3. GPT as models of brain mechanisms: lack of language productivity

Next to the boundary conditions discussed above, the question is also whether deep learning methods as found in GPT are adequate to model the productivity of human neural language processing.

For example, Ouyang et al. (2022) extended the work on language performance with GPT-3 (Brown et al., 2020) by fine-tuning the pre-trained model using human feedback. The behavior of the new model improved considerably over the old one, but it nevertheless still produced mistakes that we would classify as 'odd'.

One of these was initiated by the prompt "Why is it important to eat socks after meditating?". The old GPT-3 model just replied with "What can you learn from socks?". Instead, the new model responded with a lengthy reply. At face value, it is remarkably coherent and fluent. But it simply missed the point that socks are inedible. For example, it argued that "Some experts believe that the act of eating a sock helps the brain to come out of its altered state as a result of meditation ..." (Ouyang et al., 2022, p. 16).

The issue here is not whether the behavior of the model could be improved by more learning. Probably a model could learn that socks are inedible if suitable training material for that was provided. But this does not solve the underlying problem. Training material for any system will always be limited and most likely biased (e.g., Niven, & Kao, 2019; Bender et al., 2021). Hence, there will always be gaps in learned cognition.

Systems like GPT-3 aim to circumvent these gaps by recombining learned material in a stochastic manner. However, this will be influenced by the underlying distributions of the training data (e.g., McCoy et al., 2019). Moreover, GPT-like models are pre-trained with data sets in the order of $10^{11}$ to $10^{12}$ tokens (Brown et al., 2020). As noted, this is more than a human will see in a lifetime. But it is minute compared to the 'performance set' of $10^{20}$ sentences or more that humans would be able to process (Miller, 1967). This set is based on the human lexicon, around 60,000 words or more (Bloom, 2000), and a limited sentence length of 20 words or less. For any of these sentences humans would be able to answer 'Who does what to whom' questions, expressing the specific relations between the words in a sentence.

It is doubtful whether deep learning models like GPT models would be able to cover and express all possible relations found in this performance set based on the minute sample of this set they are trained with (even ignoring the difference between tokens and words here). For example, GPT-like models would be able to grasp that *astronaut* is the agent in the phrase *astronaut rides horse*, because it could be derived statistically from learned relations between, say, *astronaut is a human* and *human rides horse*, found in its training set. However, the reverse relation of *horse riding human* would be more difficult to derive from the training set. Indeed, GPT-like models find it much more harder to grasp the relations in *horse rides astronaut* (Marcus, 2022).

One could argue that *horse rides astronaut* does not occur in real life; therefore it is not a relevant example for understanding language processing. But this argument misrepresents human cognition. First of all, we are able to identify *horse* as the agent in this phrase. Indeed, that is why we know that it is odd. Secondly, a skilled artist would be able to draw a picture expressing the odd relations in this sentence.

Furthermore, and most importantly, we are confronted with examples of this kind. They are found extensively in cartoons, comics, games, books and movies. Young children have no difficulty in grasping the relation between a rectangular sponge and its square pants, and they can grasp that its home is a pineapple at the bottom of the see. None of this directly relates to their real-life environment. But it does so indirectly; agents can live in homes and they can wear pants. Children grasp these relations directly, even when they are filled in by odd examples, as illustrated with *SpongeBob SquarePants*. There is no need for retraining here. Moreover, the fact that young children can do this (and are in fact a main target for such cartoons) also illustrates that this ability is not based on learning massive amounts of data. Instead, it is based on learning limited amounts of data and the ability to grasp relations in a direct and explicit way.

The example of *SpongeBob* again illustrates the fundamental difference between deep learning procedures and human learning. To reiterate, children do not learn language by digesting a massive amount of examples at once. Instead, they develop language in an incremental manner, starting with two or three word sentences and progressing from that (e.g., Saxton, 2010). Moreover, when they have learned something, they can expand on that by learning new material without having to relearn the previously learned material.

I posit that the underlying ability here is 'productive cognition', by which learned relations can be applied to other instances in a productive way. So, children will learn what an agent is and what it can do in certain circumstances (e.g., *Mary lives in a home*). In this way, they are capable of recognizing aspects of agenthood in a cartoon character and understand how it would act in its environment, for example that it lives in a house. That is, they will have achieved an understanding of the abstract relation 'agent lives in house', which they can fill in with specific characters, objects and environments. They would also be able to recognize the difference between a fictional world and the real world they live in.

Productive cognition is also suitable for explicit instruction. Children could be told 'X is an agent' and 'Y is a house', even with the understanding that these relations are restricted to a specific (e.g., a cartoon) world. They could also be told that socks are inedible, if they had not experienced that already. It is very difficult to see how cognitive abilities such as these could be derived from statistical learning and stochastic recombination only. Not only would this have to cover all possible relations between e.g., agents, objects and actions in real life. But also all those that are not possible in normal circumstances, but could occur in fictional worlds, with the ability to tell the difference.

But if productive cognition is important, this raises the questions of what it is about and how it could be achieved in a neural manner. The next section discusses the first question and argues that productive cognition depends on the 'logistics of access' needed for compositional processing, as illustrated with the examples above.

## 4. Productive cognition: logistics of access

A key aspect of productive cognition is compositionality. For example, in this way the phrase *Mary lives in a home* is represented and processed as a composition of the words in the phrase in line with its relational structure. Then, it is straightforward to replace *Mary* and *home* with *SpongeBob* and *pineapple*, even if these words are newly learned. There is no need to integrate the new phrase *SpongeBob lives in a pineapple* in the training material or hope for sufficient statistical overlap with the already learned material so that this phrase can be grasped as well.

This close relation between productively and compositionality[2] has been a foundational feature of the cognitive theory known as 'classical cognition', as proposed by e.g. Fodor and Pylyshyn (1988) and Newell (1990). Classical cognition is often identified with 'symbol manipulation', because that is seen as the key operation needed for productivity. In turn, this has resulted in a continuing debate on whether symbol manipulation is essential for cognition (e.g., Marcus, 2020; Browning & LeCun, 2022).

However, the use of symbols in classical cognition is in fact derived from a specific form of 'logistics of access', which is a fundamental requirement for productive cognition. For example, in the phrase *SpongeBob lives in a pineapple* logistics of access is needed to access the representations of the words in the phrase and somehow bring them together in line with the relational structure of the phrase.

Newell (1990 p. 75) described logistics of access as a consequence of:
> "... the basic proposition behind information theory, namely, for a given technology there is a limit to the amount of encoding that can occupy a given region of physical space. (...) Thus, as the demands for variety increase (...) the local capacity will be exceeded and distal access will be required to variety that exits elsewhere."

In the view of Newell (1990), logistics of access requires the use of symbols:
> "The symbol token is the device in the medium that determines where to go outside the local region to obtain more structure. The process has two phases: first, the opening of *access* to the distal structure that is needed; and second, the *retrieval* (transport) of that structure from its distal location to the local site, so it can actually affect the processing. (...) Thus, when processing "The cat is on the mat" (...) the local computation at some point encounters "cat"; it must go from "cat" to a body of (encoded) knowledge associated with "cat" and bring back something that represents that a cat is being referred to, that the word "cat" is a noun (...) and so on. Exactly what knowledge is retrieved and how it is organized depend on the processing scheme. In all events, the structure of the token "cat" does not contain all the needed knowledge. It is elsewhere and must be accessed and retrieved." (p. 74, italics by Newell).

So, in this view logistics of access consists of distal access and retrieval, which requires the use of symbols to identify the required information and to represent and copy that information to be transported and processed (manipulated) at a 'local' site. As a result, the processing of relational structures such as *cat is on the mat* is based on symbol manipulation. This raises two questions:
1. Is the use of symbols indeed necessary for logistics of access?
2. If not, would symbols and symbol manipulation then still be needed to represent and process relational structures?

It seems that the last question has been answered negatively by deep learning. For example, the GPT model of Ouyang et al. (2022, p 16) produced "Some experts believe that the act of eating a sock helps the brain to come out of its altered state as a result of meditation ..." as an answer on whether it is important to eat socks after mediation. Regardless of whether the answer makes sense, it is remarkably structured. Indeed, many examples clearly show that models such a GPT can produce structural expressions without relying on symbols.

But this in itself does not yet answer the first question. It is distinctly possible that although GPT models can produce structural expressions, they do not possess logistics of access needed for productivity. The lack of productivity in GPT language processing discussed in section 3 indeed indicates that they do not. Hence, it would still be possible that symbols and symbol manipulation are needed for productivity.

To put this in perspective, consider a phrase like *sleepy horse rides astronaut*. With a lexicon of 60.000 words or more one could assume that the number of adjectives, nouns and verbs is each in the order of $10^4$. So, around $10^{16}$ specific phases of this kind can be produced (with intransitive verbs one could use prepositions as in *sleepy SpongeBob lives in pineapple*). Humans would be able to identify the relations in each of these sentences, whether or not they make sense. For example, they would be able to identify that in this phrase the horse is sleepy, not the astronaut. Yet, this set is already substantially larger than the pre-training sets of GPT models (Brown et al., 2020).

Of course, it is possible and indeed likely that GPT models can represent a large number of these phrases based on the statistical relations derived from training. But this does not imply that they will be able to to do that for all, if only because the set of possibilities for even this simple phrase already exceeds their training set by far.

The fact that humans are able to identify the relations in each of these sentences, even though their training set is much smaller than those of GPT like models, suggests that it would be better to integrate learning with an architecture that is designed to possess logistics of access, instead of aiming to achieve productivity based on learning alone.

Yet, one could argue that any approach based on learning, such as GPT, is always more preferable than an approach based on design. However, this argument misses the point that learning architectures are based on design as well. For example, BERT is bidirectional (Devlin et al, 2019) whereas GPT is unidirectional (Radford et al., 2018). This difference does not result from learning. That is, it is not the case that BERT and GPT are based on the same basis architecture but then develop differently as architectures based on some form of learning. Instead, they are different by design, e.g., with the aim to produce specific learning behavior or performance.

With models like these, the hope or expectation is that the chosen combination of designed architecture, learning method and training data will produce the behavior or processing capabilities aimed for. But it is not clear in advance that this will be the case. Moreover, even after some initial success it is still possible that the chosen combination could not produce the desired result after all. The failed attempts to teach natural language to chimpanzees or bonobos are again a case in point. Event the combination of their highly evolved neural systems and extensive language training from birth is apparently not sufficient to learn language in full (Terrace, 2019).

So, the choice here is not whether to start with a designed architecture or not, as any learning method starts with a designed architecture. The choice is whether or not this architecture is designed to possess logistics of access.

In theory, it is possible to start with architectures that are not designed for that, but that would be suitable to achieve it on the basis of learning. After all, humans do possess logistics of access and they learn language. Moreover, even if language development also depends on genetic factors, these have derived from an evolutionary development, which could also be seen as a learning process.

But the difficulties for such an approach are staggering, as it would depend on finding a successful combination of a suitable beginning network structure, learning rules and training data. The search for such a combination could be open-ended.

Therefore, the next sections discuss architectures that are designed to possess logistics of access. Language processing with these architectures could also answer the question of whether logistics of access always requires the use of symbols and thus implies processing in the form of symbol manipulation.

### 5. Classical architecture for logistics of access

Logistics of access in classical cognition is based on architectures like the Turing Machine, the Von Neumann Architecture (VNA) of the digital computer, or variants thereof (e.g., see Fodor and Pylyshyn, 1988; Newell, 1990). Here, I will use the VNA as the example for all of these.

Of course, the VNA is a computer. But the actual computing part of it consists of hard-wired circuits in the CPU, similar to those in a pocket calculator. What makes the VNA stand out is its bus-oriented architecture that provides logistics of access for arbitrary computations[3]. This is needed because the computing circuits are by necessity hard-wired, so their input and output registers are limited in capacity. Hence, to make computation more productive, partial results need to be copied from these registers and stored elsewhere, and retrieved back to them when needed. The memory needed for this is connected to the data-bus, with controlled access. Continuing in this manner, a whole range of memories can be connected to the data-bus (up to the internet). However, these memories are all computationally inert; computing is done only in any of the hard-wired circuits in the architecture[4].

From Newell's quotes presented in section 4, it is clear how a VNA would provide the logistics of access he described, consisting of distal access and retrieval. Because this form of logistics of access depends on the use of symbols, it is also clear why the cognitive architectures he aimed for are based on symbol manipulation.

Perhaps the reason of why Newell described logistics of access in terms of distal access and retrieval, for which symbols are necessary, resulted from the fact that it fits with the VNA, and the VNA was the only architecture of which it was known how it possessed logistics of access[5].

The need for distal access derives from the limited amount of information that can be stored physically at a given site, so it is fundamental for any productive architecture.

But the use of symbols for the identification and retrieval of distal information could result from using the VNA as the architecture for logistics of access, and could be specific for it.

Hence, logistics of access with VNA-like architectures and the use of symbol manipulation in cognitive processing are closely related. Yet, this relation of often ignored, in particular because the importance of logistics of access is overlooked.

An example is found in the neural model of sentence representation by Kriete et al. (2013). Here, dedicated local memory slots represent the agent, action and theme of a noun-verb-noun sentence. These slots can be filled with symbols in the form of neural codes (or neural symbols, for short) that represent the address codes of other local memory slots, which in turn represent information about the words involved. In this way, a noun verb noun sentence can be represented with a clear relational structure, as the role of a neural symbol in the sentence is given by the register in which it is stored. In this way, *astronaut rides horse* can be distinguished from *horse rides astronaut*.

However, this process should work for all specific noun-verb-noun sentences possible in language, even those that would be meaningless in real life but could be meaningful in a fictional world. Hence, it is necessary that there are 'links' that can be used to access and retrieve any relevant neural symbol and store it in the required local register. For example, because any noun could occur in a noun-verb relation, there would have to be a link between any neural symbol representing a noun, stored somewhere in the architecture, and the local registers involved in processing the noun-verb relation.

This suggests that a neural model of this kind would be based on a yet unknown neural variant of the VNA, in line with the conclusion of Fodor and Pylyshyn (1988) that the brain does have the structure of some kind of VNA. However, Kriete et al. (2013) do not state this explicitly, as the model does not deal with logistics of access.

A similar lack of logistics of access is found in approaches that aim to achieve symbol manipulation with neural networks (e.g., see Garcez & Lamb, 2020) or in 'binding' models that aim to represent relational structures without the use of symbols or symbol manipulation (e.g., Eliasmith, 2015; Martin & Doumas, 2019; Fitz, et al., 2020; Müller et al., 2020).

For example, Müller et al. (2020) simulate the attribution of semantic roles, e.g., the role *agent* to the word *truck*, using neural plasticity that interconnects a set of neurons (or cell assembly) that presents the word *truck* with a cell assembly that resents the agent role. Cell assemblies for other nouns (e.g., *ball*) could be connected to the agent cell assembly in a similar manner. In turn, the cell assemblies for agent, verb and theme could then be connected to represent noun verb noun structures without the use of symbols and symbol manipulation.

However, logistics of access would require the cell assemblies of all words to be connected to all cell assemblies of the potential roles the words could have, including for new words just heard (Feldman, 2013). Moreover, not just words but entire phrases could be, e.g., the agent in other phrases, as in *The reporter that the senator attacked admitted the error* (Just et al., 1996). Here, *reporter* is even simultaneously the agent in one and the object (theme) in another phrase. It is not clear how these and other issues (e.g., Jackendoff, 2002) could be resolved with a binding mechanism as proposed by Müller et al. (2020). As long as these issues are not resolved it is not clear whether such a mechanism could provide language productivity without relying on symbols and symbol manipulation.

However, as briefly described below, logistics of access can be achieved in a neural architecture. Structural relations in phrases and sentences in the architecture are represented and processed without symbol manipulation, which, e.g., results in specific predictions on brain activity in sentence processing (van der Velde, 2022).

### 6. Neural architecture for logistics of access
An example of logistics of access in a neural architecture is provided by the Head-Dependent Neural Blackboard Architecture (HD-NBA), which derived from the Neural Blackboard Architecture (van der Velde & de Kamps, 2006). The HD-NBA can represent and process any sentence structure in English, e.g., as described in Huddleston and Pullum (2002), without relying on symbol manipulation. Here, only a brief outline is presented, with an emphasis on logistics of access without the use of symbols. An extensive discussion with simulations is presented in van der Velde (2022).

The HD-NBA is inspired by a quote from William James (1890, p. 108, italics by James):
> "For the entire nervous system *is* nothing but a system of paths between a sensory *terminus a quo* and a muscular, glandular, or other *terminus ad quem*."

In the HD-NBA, arbitrary sentence structures are indeed represented and processed as 'connection paths', based on the following key ideas:
- Logistics of access, hence productivity, is solved by using small-world like network structures (e.g., Shanahan, 2010 that afford processing and creating arbitrary relational structures.
- The small-world like network structures, referred to as 'neural blackboards', interconnect all entities (e.g. 'words') that can occur in the relation.
- Logistics of access does not depend on using symbols for retrieval. Instead, the entities in the relation are each encoded in specific neural structures, which are connection paths themselves (van der Velde, 2015b). They remain 'in situ' during processing.
- Distal access between these entities is achieved by creating a connection path in the small word network structure that interconnects them. In the HD-NBA, these paths are temporal.
- Because word (concept) information always remains in situ, it is always content addressable, even when a word is part of a larger compositional structure such as a sentence.
- Relations between e.g. words in a sentence structure are encoded in a compositional manner by means of the connection paths in the neural blackboards interconnecting these words. The connection paths encode the intrinsic structure of a sentence or relation.
- The processing involved concerns controlling the flow of activation, both during the creation of these connection paths and in the process of retrieving information from it (see van der Velde & de Kamps, 2010).

A schematic illustration of some of these key ideas is given in Figure 1. The ovals represent neural representations of 'conceptual' information. In terms of the hub-and-spoke theory of semantic memory (e.g., Lambon-Ralph et al., 2017), each oval consist of a selective network structure distributed over different modal and amodal parts of the cortex. In this way, it involves one or more learned connection paths in the sensorimotor loop (van der Velde, 2015b).

Panel (a) represents a way to encode relations between concept networks, using 'conditional' connections that selectively interconnect the concept networks involved in the relation. In its basic form, a conditional connection consists of a gating circuit that can be opened (or activated) by a specific instruction (e.g., see van der Velde & de Kamps, 2006). These gates are referred to as control gates, because they are opened by an external control instruction or signal. The kind of instruction used to open a control gate determines the relation involved. Here, *cat* is connected to *pur* and *paw* with conditional connections, activated with the instructions *can* and *has*, respectively. The query *cat has?* would activate *cat* and open the *has* gate between *cat* and *paw*, producing the answer *paw*.

Panel (b) illustrates a more generic use of gating circuits, reflecting the action abilities of a cat. The generic query *cat do?* would activate any action that a cat could perform (and that is stored in memory). However, this scheme is not suitable to express what the cat is currently doing, e.g., running or eating, because both of them will be activated by the query *cat do?*.

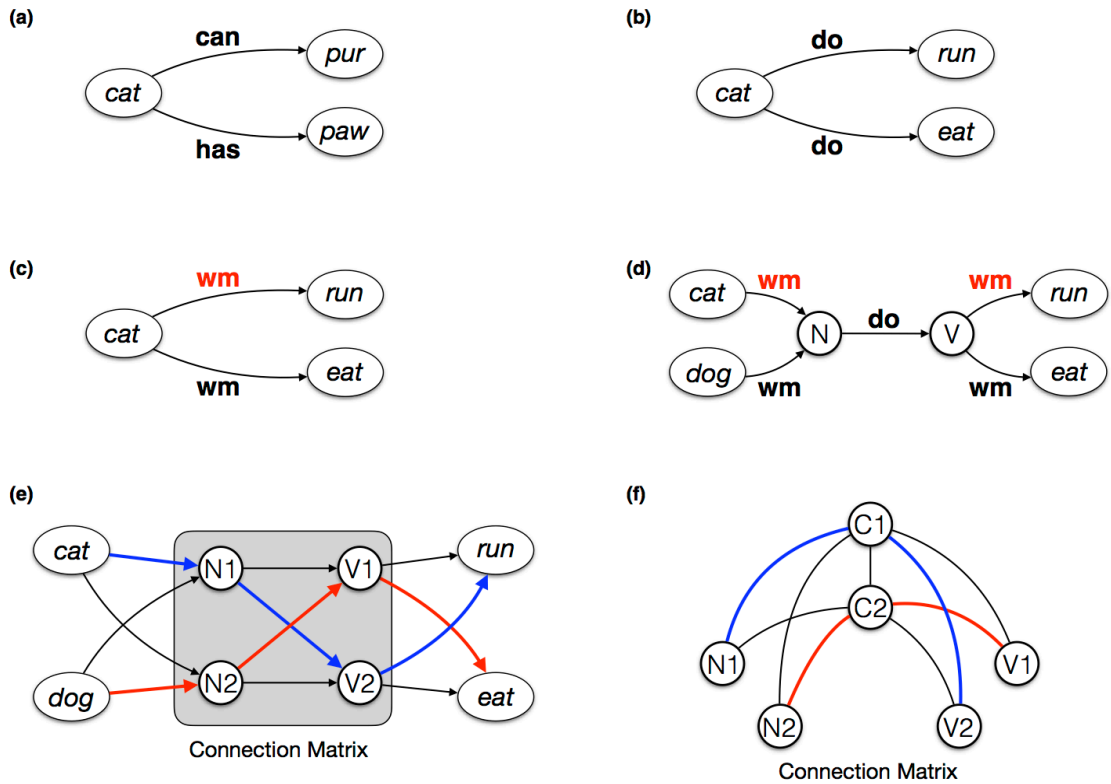

**Figure 1. Illustration of relational structures as connection paths.** Ovals represent concepts (words). Circles represent local neural populations. Arrows and links represent conditional (gated) connections. Labels next to them represent instructions to open control gates (external control) or working memory (wm) to open 'binding' gates (red is 'active').

Panel (c) illustrates how this selectivity could be achieved. Here, the conditional connections are selectively opened by an active 'working memory' (WM). This would be a local population of neurons that remains active for a while once it has been activated, as found in working memory task in the brain (e.g., Bastos et al., 2018). The specific WM population between *cat* and *run* could be activated by, say, first seeing a running cat, or hearing the sentence *cat runs*. While this WM is active, it opens the gate between *cat* and *run*, reflecting the current knowledge that a cat is running.

A WM gate is effectively a form of 'binding': as long as it is active, activation can freely flow through the gate it controls, thereby effectively binding the connected neural structures into a temporal unity. So, in panel (c) *cat* and *run* are bound. So, when *cat* is active it will activate *run*. This activation is selective, but also immediate.

Panel (d) illustrates a further extension by introducing two local neural populations, one for noun (N) and one for verbs (V). They are interconnected by a conditional connection activated by the external signal *do*, as in panel (b). The nouns *cat* and *dog* are connected to the N population and the V population is connected to *run* and *eat*, using WM gates. A current situation like *cat runs* can be encoded by activating the WM populations between *cat* and *run* and the N and V populations respectively. This could be achieved based on pattern recognition networks that classify *cat* as a noun and *run* as a verb. However, when *cat* is now active it will not immediately activate *run* as in panel (c), because the N and V populations are conditionally interconnected. Instead, only the full query *cat do?* will activate *cat* and open the gate *do*, resulting in the activation of *run* (and *run* only). Hence, the combination of binding gates, activated by WM populations, and control gates, activated by external control signals, enhances the selectivity of the flow of activation in the connection paths. A reverse connection scheme would be able to answer queries like *who do run?*.

Panel (d) also illustrates the first use of a small-world like network structure or neural blackboard. Instead of direct connections between nouns and verbs, the connection paths between them run via the N and V populations. They act like 'hubs' in the overall connection structure, significantly reducing the number of connections needed. In this basic scheme there will be many 'local' connections between all nouns and the N population and all verbs and the V population, but limited connections between these hub themselves, in line with the nature of a small world network (Watts & Strogatz, 1998).

Panel (e) elaborates this scheme. Here, there are a number of N and V populations, each of them connected to all nouns and verbs respectively. All N populations are also connected to all V populations (here illustrated in the direction N to V). A combination of binding and control gates can be used to encode multiple relations simultaneously, such as *cat runs* (blue connections) and *dog eats* (red connections). See van der Velde (2022) for more detailed and complex examples, incorporating all possible relations in sentence structures. However, the number of connections between the N and V populations is still small compared to the number of relations between nouns and verbs that can be expressed. Furthermore, the number of connections between the nouns (verbs) and the N (V) populations will be further reduced by taking the (e.g., phonetic) structure of words into account (van der Velde, 2022).

The connection structure between N and V populations in panel (e) is referred to as a 'connection matrix', because it can be used to connect each noun to each verb. This in particular makes the entire connection structure productive: any noun-verb relation can be encoded as a connection path in the overall structure, even relations never been seen before.

In Figure 1, the N and V populations together with the connection matrix form the neural blackboard of the architecture. It limits the overall connection structure because the connection matrix interconnects the hub-like N and V populations, not the nouns and verbs directly. Furthermore, a new noun or verb could be connected to the overall connection structure just by connecting it to the N or V populations, or by using its phonetic structure to do that.

Panel (f) illustrates that the structure of the connection matrix can be further elaborated. For example, by introducing populations that relate to clauses (C). In this way, sentences with hierarchical structures can be represented and processed as well.

Figure 1 provides only some basic examples of how logistics of access can be achieved in architectures based on small world network structures. However, as noted, the HD-NBA can represent and process any sentence structure in English, and it uses only one connection matrix for all word and relation types in these sentences (van der Velde, 2022).

The productivity of the neural architecture outlined above depends on a 'fixed' network structure (obtained after its development). Of course, short-term modifications of connections could occur. But language processing, for example, proceeds in terms of a rate of about 3 to 4 word per seconds (e.g., Rayner & Clifton, 2009). It is difficult to see how entirely new connections could develop within that time frame that would allow the encoding of a relation that could not be encoded by the architecture beforehand.

Logistics of access in the neural architecture depends on the use of a connection matrix. This plays the role of a data bus in the VNA, by providing the ability to form connection paths for any structural combination required (e.g., any sentence structure). But in this way, all concepts always remain 'in situ', hence content addressable, in any sentence structure. For example, the 'concept' neural networks (van der Velde, 2015b) for *cat* and *run* can be activated directly in the relation *cat runs*, as illustrated in Figure 1e.

Binding in a connection matrix could be seen as a form of 'tensor' binding (Smolensky, 1990) because a connection matrix can be seen as a (rank-2) tensor. However, concepts (words) are not bound here directly as vectors in tensors of higher ranks or as reduced vectors in other vectors (e.g., Plate, 1995). Instead, words are bound by means of their type information (e.g., noun, verb) in a connection path.

The representation and processing of a sentence structure is compositional, by interconnecting neural word structures in a connection path that encodes the relations between the words in the sentence. But the word structures remain in situ, hence content addressable, throughout. This is a main difference with symbolic accounts of sentence structures (e.g., Newell, 1990), but also with accounts of sentence structures in which words as vectors are embedded in tensors (e.g., Smolensky, 1990) or in reduced vectors (e.g., Plate, 1995; Eliasmith, 2015). In these cases, content addressability of words is lost when they are part of a sentence structure.

As Newell (1990) noted, and quoted in section 4, local processing capacity is always limited. That is also the case in the neural architecture described here. So, as with Newell's classical architecture, logistics of access in this neural architecture consists of obtaining distal access as well. But unlike a classical architecture, it does not consist of the identification and retrieval of information by using symbols. Instead, information remains in situ.

In this way, there would be 'distant' networks (e.g., from the perspective of a connection matrix) that classify a word as, say, a noun. Long-range connections from these networks would affect the binding in the connection matrix with which nouns bind to N populations (see van der Velde & de Kamps, 2010). So, there is no need to transfer information from a distant site to a local site to influence processing, as Newell argued. It suffices to create a temporal connection path between a 'distal' and 'local' site.

For Newell, the use of symbols to obtain distal access formed the basis of cognitive processing with symbol manipulation. In contrast, when concept representations remain in situ, processing does not consist of symbol manipulation. Instead, it consists of activating networks that create connection paths to encode relations and that control the flow of activation in the architecture.

For example, processing a sentence in the HD-NBA consists of creating a temporal connection path, controlled by the dependency relations (e.g., Nivre, 2008) in the sentence. The resulting connection path provides an intrinsic structure of the sentence (van der Velde, 2022).

Hence, the neural architecture outlined above effectively answers the two questions stated in section 4:
1. No, the use of symbols is not necessary for logistics of access.
2. No, symbols and symbol manipulation are not needed to represent and process relational structures.

### 6.1. Simulation of brain activity with mechanisms of neural processing

The HD-NBA provides explicit neural mechanisms for the representation and processing of sentence structures. This allows the simulation of brain activity observed in processing in a direct way.

A direct way to simulate brain activity would be to model brain activity as it is observed in a specific task or setting. An example is given by the intracranial brain activity observed during the processing of constituent phrases such as *Ten sad students* and *Ten sad students of Bill Gates*, presented in Figure 1F of Nelson et al. (2017). The activity increased during the processing of a constituent and declined after its completion, in accordance with the notion that constituents are formed almost directly (e.g., Tanenhaus et al., 1995). With the phrase *Ten sad students of Bill Gates* this rise and decline of activity was observed both with the phrase *Ten sad students* and with the entire phrase, which suggests that *Ten sad students* is processed as a partial constituent of the entirely phrase. A direct simulation of this pattern of activity based on the HD-NBA is presented in van der Velde (2022, Figure 3).

The fact that logistics of access in the HD-NBA does not depend on a data bus and symbols also affects the activity generated by the architecture. Data is not stored in registers, so there are no address codes to locate where data is stored. Therefore, control of processing operates over the entire architecture, which results in different accounts and predictions of brain activity (van der Velde, 2022, e.g. Figures 4d, 5d).

## 7. Learned cognition versus productive cognition

This section briefly discusses how neural architectures designed for logistics of access could be combined with forms of deep learning. The first relation is already clear with control of sentence processing in the HD-NBA. This could be achieved with neural networks that learn to recognize, e.g., dependency relations in sentence structures (van der Velde & de Kamps, 2010; Andor et al., 2016), which in turn control the building of a sentence structure in the architecture (van der Velde, 2022).

In a more distant way these neural architectures could be a target for processes that aim to develop them on the basis of more general or simplified architectures and mechanisms, in interaction with particular learning or development procedures. Because the logistics of access is ensured in these architectures, using them as targets could inform the study of training more basic architectures to achieve logistics of access as well.

But in a more general way, deep learning and neural architectures for logistics of access could relate to two different forms of processing that would interact in cognitive processing. As noted in the introduction, deep learning has shown that structural forms of knowledge can be embedded in and retrieved from neural networks. In a similar way, the connections in the brain could be modified by experience throughout life. Even though the actual forms of learning or development could be different, deep learning has shown that structural knowledge could be acquired in this way.

In the view proposed here, human-level cognition (natural and artificial) would result from interactions between 'learned cognition' (for want of a better term) and 'productive cognition' (again, for want of a better term). Learned cognition would share characteristics with deep learning in that its learning information would also be limited and biased, as it would be based on actual experiences encountered. Productive cognition could address these gaps in learning by explicit instruction or reasoning, based on neural architectures with logistics of access.

For example, based on learned cognition we could over time learn that birds can fly. Using productive cognition we can then come to the conclusion that 'birds can fly' is a valid proposition, without having to wait for more observations of the same kind. Of course, over time we could adjust this propositional knowledge based on further experience. But in this way, learned cognition feeds into productive cognition.

However, productive cognition will also feed into learned cognition. By explicit instruction we can acquire explicit knowledge about a novel domain. However, in becoming an expert in that domain, we then also acquire forms of learned cognition, based on prolonged 'hands-on' experiences in that domain. These could be hard to explain because they would rely on the same kind of processes and produce similar forms of network embeddings that give rise to learned cognition. This might explain why it was difficult for IBM's Watson to become a medical expert, because this not only requires acquiring a lot of medical facts, but also to integrate them with experience over time. So, expertise is not given by explicit instruction alone. It is clear that this process would be continuous, with productive cognition feeding into learned cognition and vice versa.

An example of the integration of learned and productive cognition could be found in creativity. It is generally assumed that an idea is creative if it is both novel and appropriate (e.g., Runco & Jaeger, 2012). In line with this, creativity would be the result of a cyclic interaction between generation and evaluation (e.g., Kleinmintz, et al., 2019), which could coincide with the interaction between learned and productive cognition. So, a deep-learning like network could generate the idea that eating socks would be good after meditation, which would indeed be a novel idea. But a reasoning process, productively integrating knowledge about socks, eating and meditation would then conclude that this idea is not very appropriate.

## 8. Conclusions

Deep learning networks in their present form cannot be seen as appropriate models of human neural cognition. They fail on fundamental boundary cognitions, such as the amount of learning required, and they do not provide logistics of access needed to achieve productive human cognition.

Logistics of access can be achieved with neural architectures. As exemplified with the HD-NBA, these architectures could process e.g. sentence structures without relying on symbols and symbol manipulation. Based on its small-world network structure, the HD-NBA can create and process relational structures in a compositional manner as connection paths, that temporally interconnect learned (semantic) knowledge.

In the view proposed here, learned and productive cognition are complementary in human-level cognition. In turn, this might require two complementary approaches to investigate the relation between brain and cognition. Approaches as given by deep learning could reveal how learned cognition could develop in the brain. In combination with that, however, I suggest that neural cognitive architectures should also be developed with the aim to account for the full productivity of human cognition, based on logistics of access.

One could object that these architectures would be designed. However, this is also true of deep learning architectures. Furthermore, by combining research on learning methods with productive architectures, the latter could e.g. function as targets for learning methods to develop them out of more basic neural structures. This could enhance the study of both natural and artificial human-level cognition.

**Notes**

1. There are examples of multilingual processing with GPT. However, in these cases information on the other language is either present in the original training set (e.g., Armengol-Estapé et al., 2021) or introduced specifically for the purpose of multilingual processing (e.g., Shliazhko et al., 2022). The issue here is whether GPT models could be trained so as to be able to process any human language, even ones that might not yet exist. Only then would a pre-training stage bring GPT on a par with the evolutionary development of human language. To the best of my knowledge, there is no evidence that this would be possible.

2. Another characteristic of productive cognition is systematicity (e.g., Fodor & Pylyshyn, 1988), which implies that if you understand the relations in *astronaut rides horse* you cannot but understand the relations in *horse rides astronaut*. In fact, productivity, compositionality and systematicity are closely related as characteristics of human cognition and they imply each other (Fodor & Pylyshyn, 1988).

3. The bus-oriented architecture of the VNA also makes it universally programmable, but that is not relevant for the discussion here on logistics of access.

4. Of course, one can have multiple cores for computing. But each core operates in the manner of the VNA, and the processing they can perform depends on the access to data they have. This, in general, makes genuine parallel computing difficult.

5. Given the fact that human cognition is productive, one would have to assume that the brain possesses logistic of access. But it is not known how it achieves that. Indeed, classical cognition assumed that the brain would have some kind of VNA structure, because that would be the only way to achieve productive cognition (Fodor & Pylyshyn, 1988).

## References

Andor, D., et al. (2016). Globally normalized transition-based neural networks. *arXiv:1603.06042v2*

Armengol-Estapé, J., et al. (2021). On the Multilingual Capabilities of Very Large-Scale English Language Models. *arXiv:2108.13349v1*

Bastos, A. M., et al. (2018). Laminar recordings in frontal cortex suggest distinct layers for maintenance and control of working memory. *Proc Natl Acad Sci USA, 115(5),* 1117-1122.

Bender, E. M., et al. (2021). On the Dangers of Stochastic Parrots: Can Language Models Be Too Big? In *Conference on Fairness, Accountability, and Transparency (FAccT '21),* Virtual Event, Canada. ACM, NewYork, NY, USA.

Blank, I., et al. (2014). A functional dissociation between language and multiple-demand systems revealed in patterns of BOLD signal fluctuations. *J. Neurophysiol. 112*, 1105–1118 (2014).

Bloom, P. (2000). *How children learn the meaning of words*. Cambridge, MA: MIT Press.

Brown, T.B., et al. (2020). Language Models are Few-Shot Learners. *arXiv:2005.14165.*

Browning, J & LeCun, Y (2022). What AI Can Tell Us About Intelligence. *Noema.* https://www.noemamag.com/what-ai-can-tell-us-about-intelligence/ (Accessed 30 September 2022).

Caucheteux, C., Gramfort, A. & King, J. R. (2022). Deep language algorithms predict semantic comprehension from brain activity. *Scientific Reports, 12*, 16327 https://doi.org/10.1038/s41598-022-20460-9

Devlin, J., et al., (2019). BERT: Pre-training of Deep Bidirectional Transformers for Language Understanding. *arXiv:1810.04805v2*

Eliasmith, C. (2015). *How to Build a Brain: A Neural Architecture for Biological Cognition.* Oxford: Oxford University Press.

Fedorenko, E., et al. (2016). Neural correlate of the construction of sentence meaning. *Proc. Natl. Acad. Sci. U.S.A. 113*, E6256–E6262.

Feldman, J. (2013). The neural binding problem(s). *Cognitive Neurodynamics. 7*, 1–11.

Fitz, H., et al. (2020). Neuronal spike-rate adaptation supports working memory in language processing. *Proc Natl Acad Sci U.S.A., 117(34):*20881-20889. doi: 10.1073/pnas.2000222117.

Fodor, J. A., & Pylyshyn, Z. W. (1988). Connectionism and cognitive architecture: A critical analysis. In S. Pinker and J. Mehler (eds.), *Connections and Symbols*. Cambridge, MA: MIT Press.

Garcez, A. D. & Lamb, L. C. (2020). Neurosymbolic AI: The 3rd Wave. *arXiv-2012.05876v2*

García-Martín, E., et al., (2019). Estimation of energy consumption in machine learning. *Journal of Parallel and Distributed Computing, 134*, 75-88, https://doi.org/10.1016/j.jpdc.2019.07.007.

Goldstein, A., et al. (2021). Thinking ahead: Prediction in context as a keystone of language in humans and machines. *bioRxiv* https://www.biorxiv.org/content/10.1101/2020.12.02.403477v4

Hauser, M. D., et al., (2002). The faculty of language: What is it, who has it, and how did it evolve? *Science, 298(5598), 1569–1579.*

Huddleston, R., & Pullum, G. K. (eds.), (2002). *The Cambridge Grammar of the*
21